\def\year{2021}\relax
\documentclass[letterpaper]{article} 
\usepackage{aaai21}  
\usepackage{times}  
\usepackage{helvet} 
\usepackage{courier}  
\usepackage[hyphens]{url}  
\usepackage{graphicx} 
\usepackage{natbib}  
\usepackage{caption} 
\usepackage{pifont}
\usepackage{latexsym}
\usepackage{balance}
\usepackage{amssymb}
\usepackage{amsmath}
\usepackage{algorithm}
\usepackage{algorithmic}
\usepackage{amsthm}
\usepackage{multirow}
\usepackage{pgffor}
\usepackage{epstopdf}
\usepackage{tikz}
\usepackage{epstopdf}
\usepackage{mathtools}
\usepackage{enumitem}
\usepackage{subfigure}
\usepackage{tabulary}
\usepackage{color}
\usepackage{flushend}
\usepackage{stfloats}
\usepackage{tcolorbox}
\usepackage[utf8]{inputenc}
\usepackage[english]{babel}
\usepackage{tabularx}
\usepackage{CJKutf8}
\usepackage{microtype}
\usepackage{textcomp}
\usepackage{booktabs}
\usepackage{colortbl}
\usepackage{soul}
\usepackage{appendix}
\usepackage[switch]{lineno}
\usepackage{xspace}

\newcommand{\mn}{KG-BART\xspace}
\newcommand{\data}{CommonGen\xspace}
\title{KG-BART: Knowledge Graph-Augmented BART for Generative \\ Commonsense Reasoning}
\author{Ye Liu$^1$, Yao Wan$^2$, Lifang He$^3$, Hao Peng$^4$, Philip S. Yu$^{1}$\\
\normalsize$^1$Department of Computer Science, University of Illinois at Chicago, Chicago, IL, USA\\
$^2$School of Computer Science and Technology, Huazhong University of Science and Technology, Wuhan, China\\
$^3$Department of Computer Science and Engineering, Lehigh University, Bethlehem, PA, USA\\
$^4$Beijing Advanced Innovation Center for Big Data and Brain Computing, Beihang University, Beijing, China\\
\{yliu279, psyu\}@uic.edu, wanyao@hust.edu.cn, lih319@lehigh.edu, penghao@act.buaa.edu.cn
}

\begin{document}
\maketitle
\begin{abstract}
Generative commonsense reasoning which aims to empower machines to generate sentences with the capacity of reasoning over a set of concepts is a critical bottleneck for text generation.
Even the state-of-the-art pre-trained language generation models struggle at this task and often produce implausible and anomalous sentences. 
One reason is that they rarely consider incorporating the knowledge graph which can provide rich relational information among the commonsense concepts. 
To promote the ability of commonsense reasoning for text generation, we propose a novel knowledge graph-augmented pre-trained language generation model {\mn}, which encompasses the complex relations of concepts through the knowledge graph and produces more logical and natural sentences as output.
Moreover, {\mn} can leverage the graph attention to aggregate the rich concept semantics that enhances the model generalization on unseen concept sets.
Experiments on benchmark {\data} dataset verify the effectiveness of our proposed approach by comparing with several strong pre-trained language generation models, particularly \mn outperforms BART by 5.80, 4.60, in terms of BLEU-3, 4.
Moreover, we also show that the generated context by our model can work as background scenarios to benefit downstream commonsense QA tasks.\footnote{Our code is available at https://github.com/yeliu918/KG-BART}
\end{abstract}

\section{Introduction}

Nowadays, numerous benchmarks for commonsense reasoning have been developed to make computers more competent and human-aware. 
In particular, various pre-trained approaches have achieved impressive performance on the \textit{discriminative} commonsense tasks -- i.e., AI systems are required to choose the correct option from a set of choices based on a given context~\cite{lincommongen}, such as CommonsenseQA~\cite{talmor2018commonsenseqa}, COSMOSQA~\cite{huang2019cosmos} and WinoGrande~\cite{sakaguchi2019winogrande}. 
However, commonsense reasoning in text generation, known as \textit{generative} commonsense reasoning, still remains a challenge to existing models, which requires machines to generate a sentence describing a day-to-day scene using concepts from a given concept set.

\begin{figure}[t]
\centering
\includegraphics[width=0.9 \linewidth]{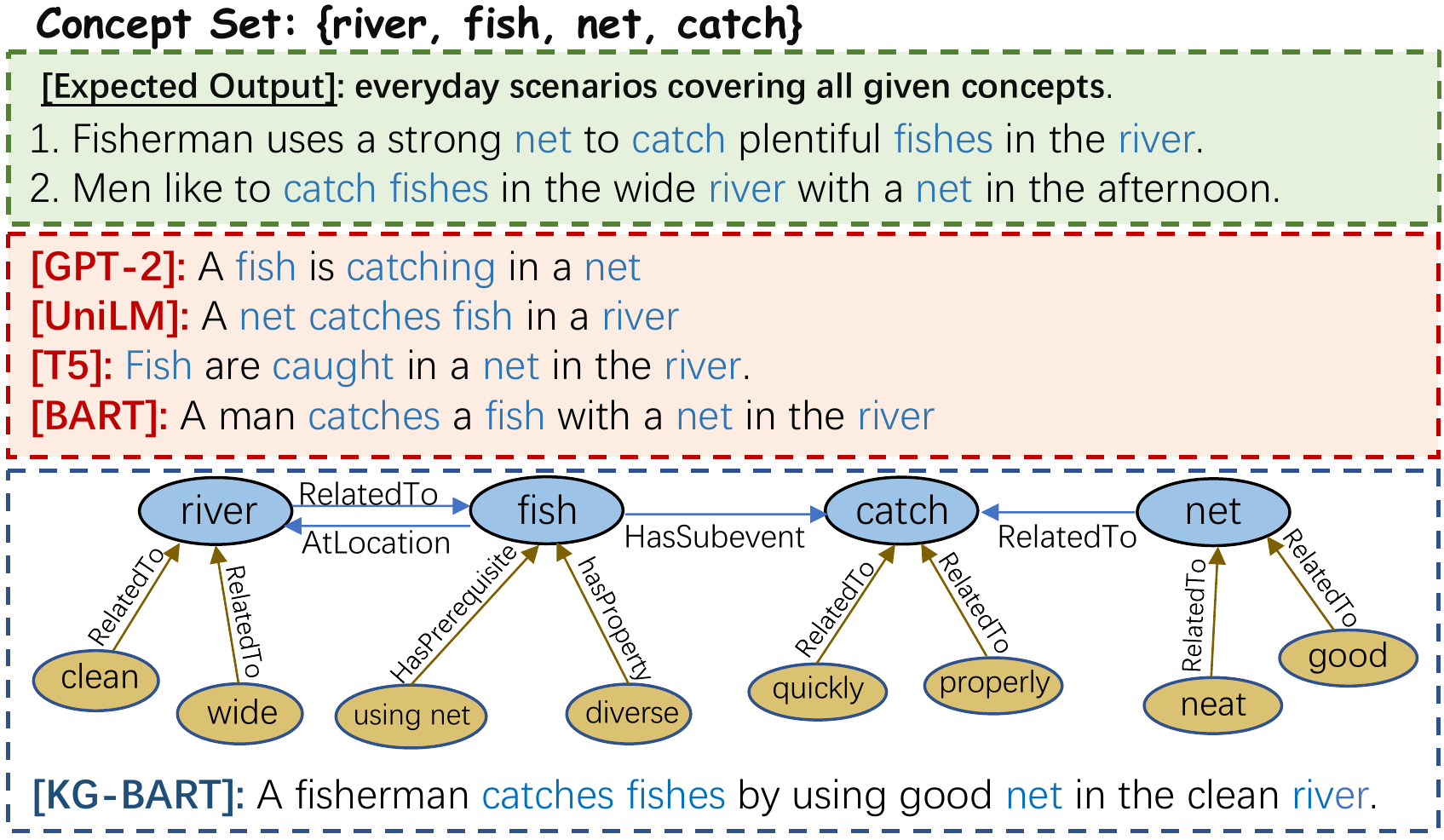}
\caption{An example of the generation outputs of our KG-BART model (blue dotted box) and the existing models without knowledge graph augmentation (red dotted box).}
\label{figure:intro}
\end{figure}

In recent years, many pre-trained language generation models have been presented for text generation tasks, such as GPTs~\cite{radford2019language,brown2020language}, UniLM~\cite{dong2019unified}, T5~\cite{raffel2019exploring} and BART~\cite{lewis2020bart}. 
Although they can capture rich language information from text sentence corpus and generate accurate language texts, almost all of them ignore knowledge information and thereby fail to generate output towards capturing the human commonsense. 
For example, as shown in Figure~\ref{figure:intro}, given a set of commonsense concepts \{\textit{river}, \textit{fish}, \textit{net}, \textit{catch}\}, the task is to generate a coherent sentence describing a scenario covering all given concepts, such as ``\textit{Fisherman uses a strong net to catch plentiful fishes in the river}''. 
From our analysis, we note that the state-of-the-art pre-trained models generate implausible and anomalous sentences in this task (red dotted box) - e.g., GPT-2 generated ``\textit{A fish is catching in a net}'', UniLM generated ``\textit{A net catches fish}'', etc. 
Moreover, the generated sentences by the pre-trained models are simple and rigid, while the human sentence is more natural and rich, like ``\textit{plentiful fishes}'', ``\textit{wide river}'', etc.

In this paper, we argue that only using pre-trained language models with textual concepts alone cannot provide sufficient information for generative commonsense reasoning. 
The commonsense knowledge graphs (KGs)~\cite{speer2016conceptnet,sap2019atomic} have been developed especially for knowledge representation in symbolic systems, and they provide a lot of candidate commonsense facts mined from corpora, which have been widely used in commonsense QA tasks~\cite{lin2019kagnet}. 
It would be beneficial to develop a model that can exploit commonsense KGs for generative commonsense reasoning task. 
For example, as shown in Figure~\ref{figure:intro}, by considering knowledge facts ``$<$\textit{fish}, \textit{HasPrerequisite}, \textit{using net}$>$'' and ``$<$\textit{fish}, \textit{HasSubevent}, \textit{catch}$>$'', it is easy to recognize the relation between concepts $\{$\textit{fish}, \textit{net}, \textit{catch}$\}$, namely using the net to catch fish. Furthermore, the commonsense relation, like ``$<$\textit{river}, \textit{RelatedTo}, \textit{clean}$>$'', can provide the adjunct word to facilitate generating a more natural and plausible daily scenario sentence. 

In light of the fact that the knowledge graph can provide the relational information to enhance the reasoning capacity and provide adjunct words to the concept, we propose a novel Knowledge Graph-Augmented framework for generative commonsense reasoning. 
It has two major steps: knowledge graph grounding and graph-based encoder-decoder modeling. 
We first construct two KGs, one is the concept-reasoning graph and another is the concept-expanding graph, both of which encode the entity representations and their dependency relations. 
Secondly, we propose an encoder-decoder neural architecture, named ({\mn}), by incorporating the grounded KGs into the state-of-the-art pre-trained language generation model BART. 
{\mn} follows the BART architecture, but instead of using the traditional Transformer, we introduce an effective Knowledge Graph-Augmented Transformer to capture the relations between concept set, where the grounded KGs are used as the additional inputs to the graph attention mechanism.
Besides, since the token and concept entity are at different granularity levels, we integrate the text representation with the knowledge concept for relational reasoning and then disintegrate to the token-level.

Overall, the main contributions of this paper are as follows:
\begin{itemize}
\item To the best of our knowledge, this is the first time that the KG is incorporated into the pre-trained model to improve the ability of commonsense reasoning in text generation.
\item We build the concept-reasoning graph to guide the pre-trained model to better reasoning the relationships among concepts. Moreover, we build the concept-expanding graph which considers both the inter-concept relation and intra-concept relation for KG-Augmented decoder to generate more natural and plausible output.
\item We propose {\mn}, a pre-trained method that is designed to better generate language via knowledge graphs and texts, and enhance the model generalization on unseen concept sets. Particularly, the integration and disintegration components are introduced to fuse the heterogeneous information between the token and concept entity. 
\item The experimental results show that {\mn} significantly outperforms the state-of-the-art pre-trained models on the task of generative commonsense reasoning. Additionally, we show that {\mn} can benefit downstream tasks (e.g., commonsense QA) via generating useful context as background scenarios. 
\end{itemize}

\section{Problem Formulation}
\textbf{Notation}.
We use $\mathcal{X}$ to denote the space of all possible concept sets, and use $\mathcal{T}$ and $\mathcal{C}$ to denote the token vocabulary and concept vocabulary, respectively.
The knowledge graph (KG) is denoted as $\mathcal{G} = (\mathcal{V}, \mathcal{E}, \mathcal{R})$, where $\mathcal{V}$ is the set of entities, $\mathcal{E}$ is the set of edges and $\mathcal{R}$ is the set of relations among entities. 
For a pair of entities $v_i \in \mathcal{V}$ (subject) and $v_j \in \mathcal{V}$ (object), associated with the relation $r_{ij} \in \mathcal{R}$, the edge $e_{ij} \in \mathcal{E}$ can be represented as a triplet $(v_i, r_{ij}, v_j)$. Specifically, we assume the concept vocabulary is a subset of KG's unigram entities, namely $\mathcal{C} \subset \mathcal{V}$.

Given an unordered set of $k$ commonsense concepts $x=\{c_{1}, c_{2}, \ldots, c_{k}\}$, where each concept $c_{i} \in \mathcal{C} \subset \mathcal{X}$ is an object (noun) or action (verb), the ultimate goal of generative commonsense reasoning is to generate a natural language output $y=\{y_{1}, y_{2}, \ldots, y_{l}\}$ that is both correct (or valid) and natural sounding for that scenario. This is often modeled by learning a function $f: \mathcal{X} \to \mathcal{Y}$ that maps the concept set $x \in \mathcal{X}$ into a sentence $y \in \mathcal{Y}$. Our aim is to boost the performance of this task with the help of KG database $\mathcal{G}$ which can be treated as auxiliary information.

More formally, we formulate the problem as follows: $h: \{\mathcal{X}, \mathcal{G} \} \to \{\mathcal{G}^R, \mathcal{G}^E \}$ that takes the concept sets $x \in \mathcal{X}$ and the knowledge $\mathcal{G}$ as the input to first learn a concept-reasoning graph $\mathcal{G}^R$ and a hierarchical concept-expanding graph $\mathcal{G}^E$, and then $g: \{ \mathcal{X}, \mathcal{G}^{R}, \mathcal{G}^{E}\} \to \mathcal{Y}$ to generate the final outputs.
Specifically, $\mathcal{G}^{R} \subset \mathcal{G}$ consisting of all concept triplets $(v^{R}_{i}, r^{R}_{ij}, v^{R}_{j})$, where $v^{R}_{i}$ and $v^{R}_{j} \in \mathcal{X}$ and $r^{R}_{ij} \in \mathcal{R}$ is the relation between each concept pairs. $\mathcal{G}^{E} = \{\mathcal{G}^{R} \cup \mathcal{N}(v^R) \}\subset \mathcal{G}$ is used to enrich the graph with adjunct information, where $\mathcal{N}({v^{R}})$ characterizes the neighborhood relationship between concept ($v^{R}$) and its adjacencies in the KG database.

\section{Knowledge Graph Grounding}
In this section, we explain how to construct and learn the embedding representations of the concept-reasoning graph and the hierarchical concept-expanding graph from the large commonsense KG Conceptnet~\cite{speer2016conceptnet}.\footnote{https://github.com/commonsense/conceptnet5/wiki/Downloads}

In the generative commonsense reasoning task, traditional pre-trained methods usually encode the concept ($x$) and decode the sentence ($y$) based on text information alone, which ignore the structural information and relations between concepts and suffer from generating a lot of implausible sentences. In order to overcome this drawback, we propose to hybridize the KG and text information in the encoder and decoder modules. Specifically, in the encoder phase, we construct a concept-reasoning graph $\mathcal{G}^{R}$ to encompass the relations between the concept set. In the decoder phase, we construct a hierarchical concept-expanding graph $\mathcal{G}^{E}$ to enrich the concept structure with the neighborhood correlation preserved in the KG database.
Based on our assumption, each concept corresponds to a KG's unigram entity, so we can directly match the concept set to the entities from KG to generate $\mathcal{G}^{R}$. In order to establish $\mathcal{G}^{E}$, we couple $\mathcal{G}^{R}$ with the association of selected neighboring nodes with each concept in KG. 
For many concepts, there are hundreds or thousands of neighboring nodes connected with each of them (via triplets) in KG, which provide us not only rich information but also less important or less relevant entities that may be undesirable. For instance, given a concept-set $\{$\textit{ski}, \textit{skier}, \textit{mountain}$\}$, considering the adjunct concepts for ``\textit{mountain}'', ``\textit{snowy}'' is more precise than others like ``\textit{small}'' or ``\textit{flat}'' according to the close semantics of ``\textit{snowy}'' and ``\textit{ski/skier}''. 
Based on this fact, we rank the neighboring nodes of each concept according to the word similarity scores and select their potential top-$k$ neighboring nodes adding to $\mathcal{G}^{R}$, so as to get $\mathcal{G}^{E}$. To calculate the word similarity scores, we use the pre-trained GloVe embedding~\cite{pennington2014glove} as the representation of each entity node in KG.
The ranking score for a particular neighboring node is the sum of similarity scores with all concepts. Here we use the cosine similarity for its simplicity and wide application.

Since some of concept pairs do not have a direct connection in the KG and some of the concept pairs connect by multiple relations, instead of directly using $\mathcal{G}^{R}$ and $\mathcal{G}^{E}$, we use a knowledge embedding method named TransE~\cite{bordes2013translating} to learn their entity and relation embeddings. 
To prepare the training triplets of TransE model, we first collect the triplets in the one-hop path, two-hop path, and three-hop path between each concept pair. Moreover, we further collect the triples between each concept node and their neighboring nodes as follows: if the concept node is the object (noun), only the neighboring node containing the adjective word will be selected; if the concept node is action (verb), only the node containing adverb word will be selected. TransE model is trained based on those selected triplets, which generates the node embedding $\mathbf{v}_i \in \mathbb{R}^{d_e}$ for each node $v_{i}$ and relation embedding $\mathbf{r}_{ij} \in \mathbb{R}^{d_r}$ for each edge $e_{ij}$. For $\mathcal{G}^{R}$, we denote each concept embedding as $\textbf{v}^{R}$, and relation embeddings as $\mathbf{r}^{R}_{ij}=\mathbf{v}^{R}_{i} - \mathbf{v}^{R}_{j}$ instead of the output of TransE to avoid missing relations between concepts.
For $\mathcal{G}^{E}$, since those neighboring nodes are connected with the concepts in the KG, we directly add their node embeddings $\mathbf{v}^{N}$ and relation embeddings $\mathbf{r}^{N}$ to $\mathcal{G}^{R}$.


\section{Graph-Based Encoder-Decoder Modeling}
\begin{figure}[t]
\centering
\includegraphics[width=1\linewidth]{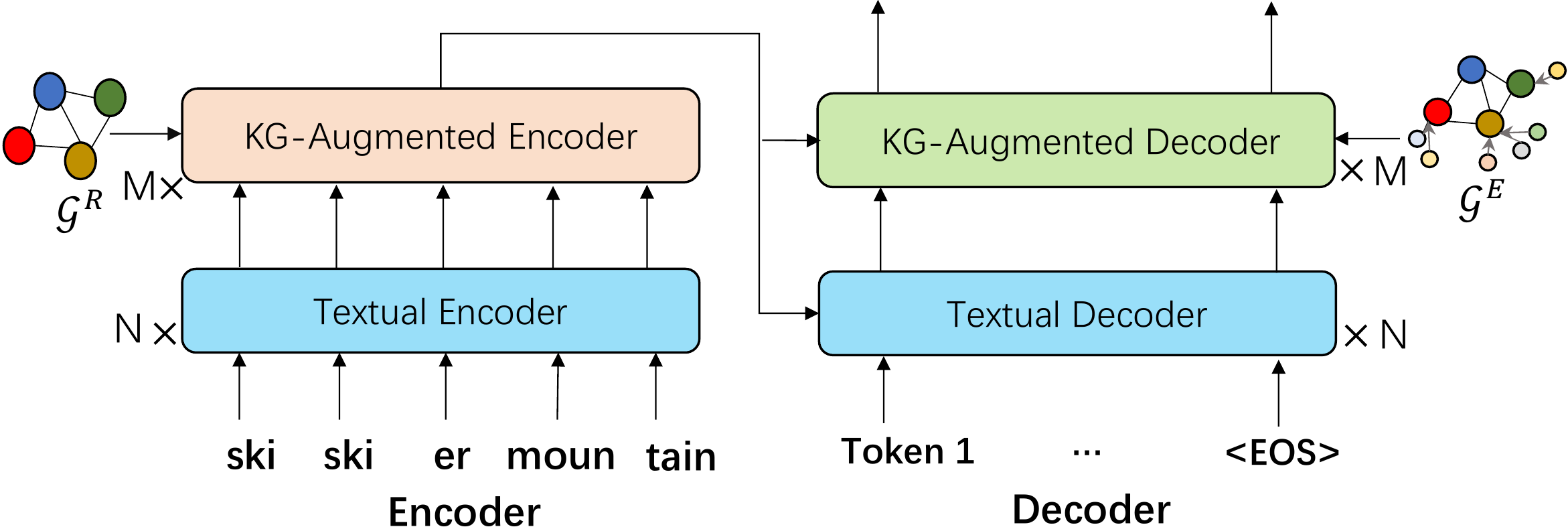}
\caption{The proposed {\mn} model.}
\label{figure:method1}
\end{figure}

\textbf{Overview}. Figure~\ref{figure:method1} presents an overview of the proposed \mn model, which follows the BART encoder-decoder architecture but uses both text concepts and KG as the input. 
The encoder is composed of two components: one traditional textual Transformer encoder module~\cite{vaswani2017attention} to represent the contextual information of each token; and another KG-augmented Transformer module based on graph attention mechanism to integrate the entity-oriented knowledge information into token representation. 
Similarly, the decoder is also composed of a stack of a textual Transformer decoder module and a KG-augmented Transformer decoder module to generate sentences with the ability of commonsense reasoning. 
Specially, we use a hierarchical graph attention mechanism to refine the KG-augmented decoder to capture the inherent structural correlations
of intra-concept and inter-concept in the graph. 
%
Note that all the node and relation embeddings are held fixed in the training process of KG-BART.
Since our textual Transformers are the same as that used in BART, here we exclude a comprehensive description of these modules and refer readers to~\cite{lewis2020bart} and~\cite{vaswani2017attention} for more details.
In the following, we will focus on the proposed KG-augmented Transformer.

\begin{figure}[t]
\centering
\includegraphics[width=0.9\linewidth]{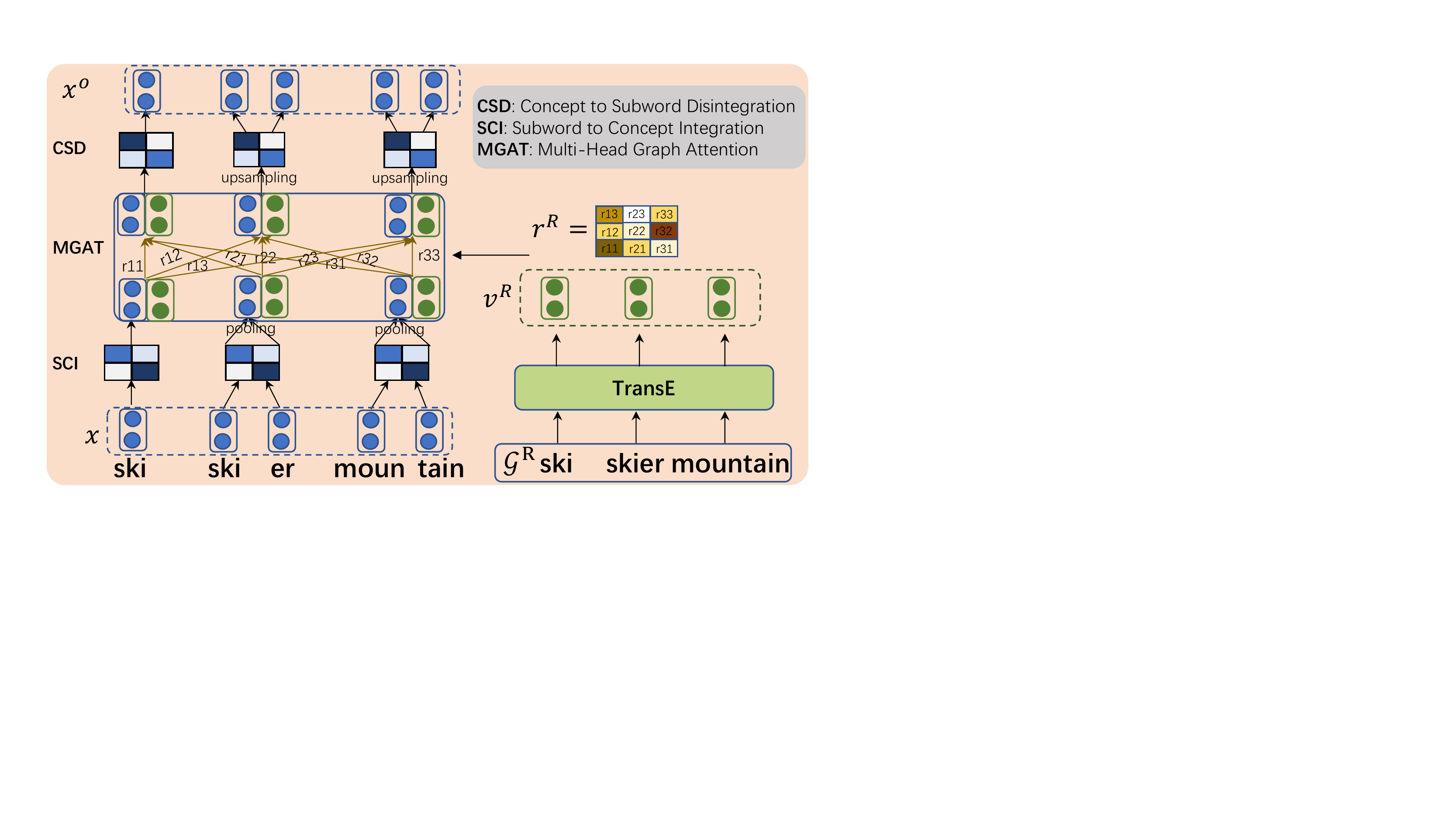}
\caption{The KG-augmented encoder.}
\label{figure:method2}
\end{figure}
\subsection{KG-Augmented Encoder} \label{sec:encoder}
As shown in Figure~\ref{figure:method2}, above the textual encoders, the KG-augmented encoder is designed to enrich the token representation by considering the KG structure. 
We propose to incorporate graph representations into the neural encoding process via a graph-informed attention mechanism. 
It takes advantage of the explicit relations to learn better intra-concept relations.
Formally, the KG-augmented encoder integrates the input token embeddings $\{\mathbf{x}_1, \ldots, \mathbf{x}_n\}$, which is the output of the textual encoders, as well as the embedding of concept-reasoning graph $\mathcal{G}^{R}$ to update the token representation as $\{\mathbf{x}^{o}_1, \ldots, \mathbf{x}^{o}_n\}$.
\subsubsection{Subword to Concept Integration (SCI)}
As the input token embeddings are based on a sequence of subwords, while our concepts in the KG are at word-level, we need to align these different granularity sequences. 
To apply the relation between concepts, we group the subwords for each concept. 
In particular, we adopt one convolutional neural network (CNN)~\cite{kim2014convolutional} with a max-pooling layer to efficiently obtain the representation in word-level.

Here we take a concrete concept as an example to better illustrate this process.
Supposing that a concept $c_{i}$ is made up of a sequence of subwords $\{x_{1}, x_{2}, \ldots, x_{m}\}$, where $m$ is the number of subwords. 
Given the token embeddings $\mathbf{x}$ from textual encoder, we first utilize a Conv1D layer, $\mathbf{x}^{\prime}_{t}=\mathbf{Z}\left(\mathbf{x}_{t}, \mathbf{x}_{t+1}, \ldots, \mathbf{x}_{t+l-1}\right)^{T}, t\in [1, m-l+1]$, where $\mathbf{Z} = [z_{1}, \ldots, z_{l}] \in \mathbb{R}^{1\times l}$ is trainable parameters and $k$ is the kernel size. 
We then apply a max-pooling layer over a sequence of the output embeddings after Conv1D:
\begin{align}
 \mathbf{e}\left(c_{i}\right)=\operatorname{MaxPooling}\left(\mathbf{x}^{\prime}_{1}, \ldots, \mathbf{x}^{\prime}_{m-l+1}\right).
\end{align}
Therefore, the final word-level textual embedding of concept is represented as $\mathbf{e}^{w} = \{\mathbf{e}(c_{1}), \ldots, \mathbf{e}(c_{l})\} \in \mathbb{R}^{k \times d_{w}}$ where $d_{w}$ denotes the dimension of concept embedding.  

\subsubsection{Multi-Head Graph Attention (MGAT)}
Given the embedding representation of concept-reasoning graph $\mathcal{G}^{R}$ with node features $\mathbf{v}^{R} \in \mathbb{R}^{k\times d_{e}}$ and relation features $\mathbf{r}^{R}$, we apply the graph attention networks (GAT)~\cite{velivckovic2017graph} to iteratively update the representations for each concept $\mathbf{v}^{R}_{i}$ through its neighbors $\mathcal{N}^{R}_{i}$.
We denote the word-level hidden state as $\mathbf{h}_{i} \in \mathbb{R}^{d_h}$, where $i \in (1, \ldots, k)$.
We further modify the GAT layer to infuse the pairwise relation embedding $\mathbf{r}^{R}_{ij} \in \mathbb{R}^{d_r}$. 
Therefore, the multi-head graph attention can be denoted as:
\begin{equation}
\small
\begin{split}
  &  \mathbf{H} = [\mathbf{e}^{w}; ~\mathbf{W}_{e}\mathbf{v}^{R}], \\
  &  z_{i j}=\operatorname{LeakyReLU}\left(\mathbf{W}_{a}\left[\mathbf{W}_{q} \mathbf{h}_{i}; \mathbf{W}_{k} \mathbf{h}_{j}; \mathbf{W}_{r}\mathbf{r}^{R}_{ij}\right]\right), \\ 
  & \alpha_{i j}=\frac{\exp \left(z_{i j}\right)}{\sum_{l=1}^{|\mathcal{N}^{R}_{i}|} \exp \left(z_{i l}\right)}, \quad
  \mathbf{h}^{\prime}_{i}=\|_{k=1}^{K} \sigma\left(\sum_{j=1}^{|\mathcal{N}^{R}_{i}|} \alpha_{i j}^{k} \mathbf{W}_{v}^{k} \mathbf{h}_{i}\right),
  \end{split}
\label{ga}
\end{equation}
where $K$ is the multi-head number, $||_{k=1}^{K}$ denotes an operation of multi-head used in Transformer, which concatenates the attention embeddings from different heads and feeds the result into a linear projection. $\mathbf{W}_{a}, \mathbf{W}_{e}, \mathbf{W}_{r}, \mathbf{W}_{q}, \mathbf{W}_{k}$ and $\mathbf{W}_{v}$ are trainable weights and $\alpha_{ij}$ is the attention weight between $\mathbf{h}_i$ and $\mathbf{h}_j$. The word-level hidden state $\textbf{H}$ contains the latent dependencies between any two concepts from textual aspect information $\mathbf{e}^{w}$ and KG aspect information $\mathbf{v}^{R}$. And $\mathbf{r}^R$ incorporates relation representations as prior constraints into the encoding process. 
In this way, our model can learn better and richer concept representations containing the relationship among concepts.

\subsubsection{Concept to Subword Disintegration (CSD)}
After updating the word-level hidden state considering the relation between concepts in the KG, we need to disintegrate the concept to the subword-level for the following process. 
We first upsample word-level hidden state $\mathbf{h}^{\prime}_{i}$ with $(m-l+1)$ times (the length before MaxPooling) as $[\mathbf{h}_{i}^{\prime 1}, \dots, \mathbf{h}_{i}^{\prime m-l+1}]$ and utilize a Deconv1D layer with vector $\mathbf{Z} = [z_{0}, \ldots, z_{l}]\in \mathbb{R}^{1 \times l}$ used in Conv1D to form the Deconv1D matrix $\mathbf{Z}_{D} \in \mathbb{R}^{m\times (m-l+1)}$ to get the subword-level hidden state $\mathbf{u}_i$: 
\begin{equation}
\small
  [\mathbf{u}_{i}^{1}, \ldots, \mathbf{u}_{i}^{m}]^{T} = \left(\begin{array}{cccc}z_{0} & & &  \\ \cdots & z_{0} & &  \\ z_{l} & \cdots & \cdots &  \\  & z_{l} &  & z_{0}  \\  &  &  & \cdots  \\  &  &  & z_{l} \end{array}\right) *\left(\begin{array}{c}\mathbf{h}_{i}^{\prime 1} \\ \mathbf{h}_{i}^{\prime 2} \\\cdot\\\cdot\\\cdot\\ \mathbf{h}_{i}^{\prime m-l+1}\end{array}\right).
\end{equation}

Then, a two-layer feed-forward network with GeLU activation~\cite{hendrycks2016gaussian} function and a residual layer normalization are applied to obtain the final output can be represented $\mathbf{x}_{i}^{o}$:
\begin{equation}
\begin{split}
    & \mathbf{p}_{i} =\mathbf{W}_{o 2} \operatorname{GeLU}\left(\mathbf{W}_{o 1}\left(\mathbf{u}_{i}+\mathbf{x}_{i}\right)\right), \\
    &\mathbf{x}^{o}_{i} =\operatorname{LayerNorm}\left(\mathbf{p}_{i}+\mathbf{x}_{i}\right),
\end{split}
\end{equation}
where $\mathbf{W}_{o 1} \in \mathbb{R}^{d_{f}\times d_{h}}$ and $\mathbf{W}_{o 2} \in \mathbb{R}^{d_{h}\times d_{f}}$ are learnable parameters, $d_{f}$ is the hidden size of the feedforward layer.

\subsection{KG-Augmented Decoder} \label{sec:decoder}
\begin{figure}[t]
\centering
\includegraphics[width=0.9\linewidth]{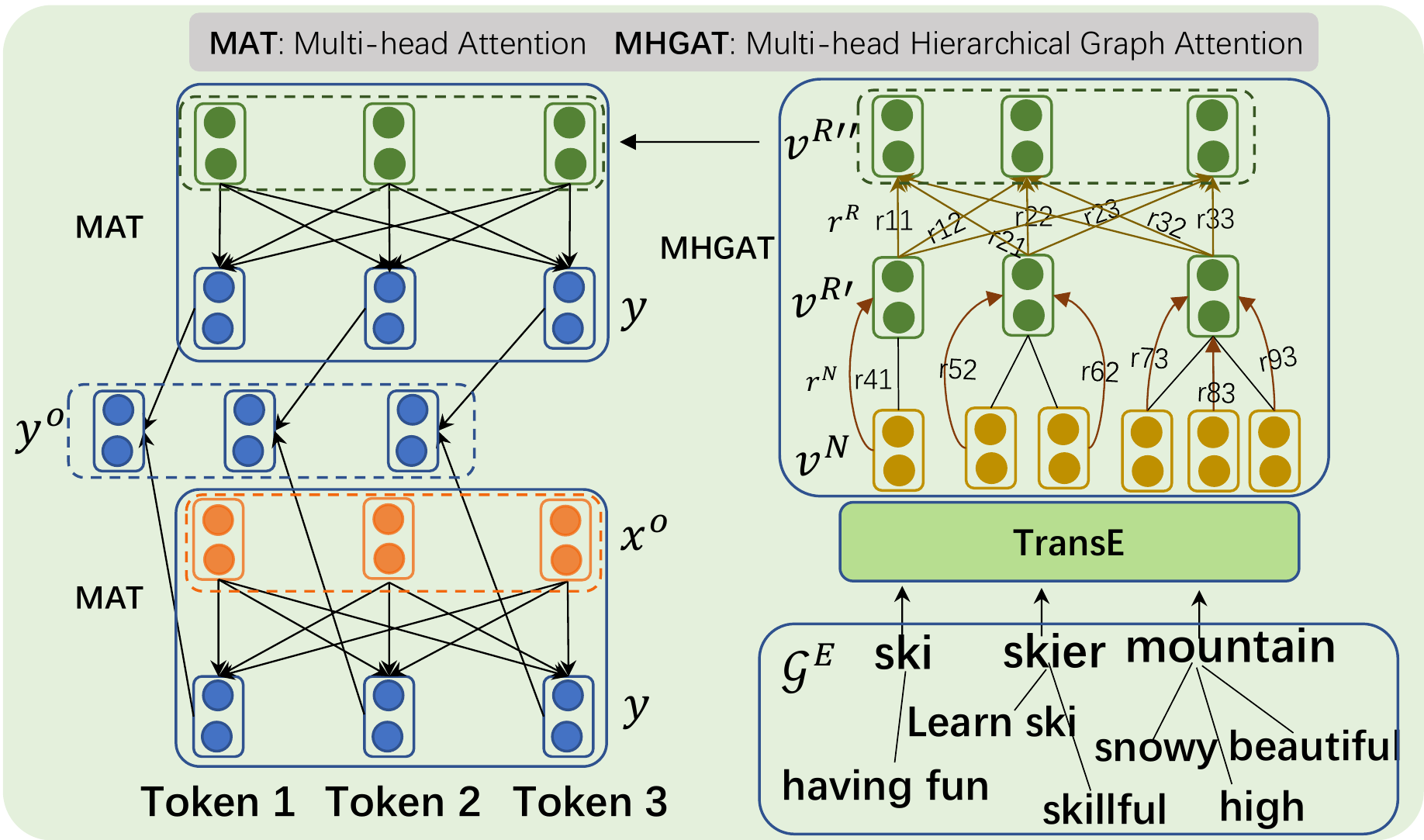}
\caption{The KG-augmented decoder.}
\label{figure:method3}
\end{figure}

In this section, our KG-augmented decoder, as shown in Figure~\ref{figure:method3}, incorporates hierarchical graph structure into the decoding process to capture the relations between concepts and their neighboring nodes which can help to generate more precise and natural output. To embody the hierarchical concept-expanding graph $\mathcal{G}^{E}$ with the generation process, we propose the multi-head hierarchical graph attention layer.  

\subsubsection{Multi-Head Hierarchical Graph Attention (MHGAT)}
To contain the adjunct description for the concept node, the first layer of hierarchical graph attention is to update the concept node $\mathbf{v}^{R}_i \in \mathbb{R}^{de}$ through its inter-concept neighboring nodes $\mathcal{N}^{N}_i$ with relation embedding $\mathbf{r}^N_{ij} \in \mathbb{R}^{dr}$. 
\begin{equation}
\small
\begin{split}
&z_{ij}=\operatorname{LeakyReLU}\left
(\mathbf{W}_{a}\left[\mathbf{W}_{q} \mathbf{v}^{R}_{i} ; \mathbf{W}_{k} \mathbf{v}^{N}_{j}; \mathbf{W}_{r} \mathbf{r}^N_{ij}\right]\right), \\
  & \alpha_{i j}=\frac{\exp \left(z_{i j}\right)}{\sum_{l=1}^{|\mathcal{N}^{N}_{i}|} \exp \left(z_{i l}\right)}, \quad
  \mathbf{v}^{R\prime}_{i}=\|_{k=1}^{K} \sigma \left(\sum_{j=1}^{|\mathcal{N}^{N}_{i}|} \alpha^{k}_{i j} \mathbf{W}^{k}_{v} \mathbf{v}^R_{j}\right). 
  \end{split}
\end{equation}

After updating the concepts with their neighboring nodes, the concepts get their new embedding $\mathbf{v}^{R\prime}$. The second graph attention layer updates the concept representation considering the intra-concept relations $\mathbf{r}^{R}_{ij} \in \mathbb{R}^{dr}$. 
\begin{equation}
\small
\begin{split}
  & z_{ij}=\operatorname{LeakyReLU}\left(\mathbf{W}_{a}\left[\mathbf{W}_{q} \mathbf{v}^{R\prime}_{i} ; \mathbf{W}_{k} \mathbf{v}^{R\prime}_{j}; \mathbf{W}_{r} \mathbf{r}^{R}_{ij}\right]\right), \\ 
  & \alpha_{i j}=\frac{\exp \left(z_{i j}\right)}{\sum_{l=1}^{|\mathcal{N}^{R}_{i}|} \exp \left(z_{i l}\right)}, \quad
  \mathbf{v}^{R\prime\prime}_{i}=\|_{k=1}^{K} \sigma\left(\sum_{j=1}^{|\mathcal{N}^{R}_{i}|} \alpha^k_{i j} \mathbf{W}^k_{v} \mathbf{v}^{R\prime}_{j}\right). 
  \end{split}
\end{equation}

We further compute the two multi-head attention (MAT)~\cite{vaswani2017attention} to capture textual and KG influence. One is the attention between the encoder hidden state $\mathbf{x}^{o}$ and the previously generated token hidden state $\mathbf{y}$. The other is the attention between the updated concept embeddings $\mathbf{v}^{R\prime\prime}$ and the previously generated token hidden state $\mathbf{y}$ as follows:
\begin{align}
    \text {AT}^{\text{KG}}=\operatorname{MAT}( \mathbf{y},\mathbf{v}^{R\prime\prime}, \mathbf{v}^{R\prime\prime}), ~~
     \text {AT}^{\text{TX}}=\operatorname{MAT}(\mathbf{y},\mathbf{x}^{o},\mathbf{x}^{o}). \nonumber
\end{align}
The final decoder output is the concatenate of the two attention with a residual connection as:
\begin{align}
    \mathbf{y}^{o} = \mathbf{W}_{att}[\text{AT}^{\text{KG}}; ~\text {AT}^{\text{TX}}] + \mathbf{y},
\end{align}
where $\mathbf{W}_{att} \in \mathbb{R}^{d_{h}\times 2d_{h}}$ is the trainable weight. $\mathbf{y}^{o}$ is used to predict the token sequence:    $P_{\text{vocab}}=\operatorname{softmax}\left(\mathbf{W}_{\text{out}}\mathbf{y}^{o} + \text{b}_{\text{out}}\right)$, $\mathbf{W}_{att} \in \mathbb{R}^{V\times d_{h}}$ and $V$ is the vocabulary size. 

\subsection{{\mn} Model Pre-Training}
The embedding vectors of words in text and nodes/entities in KG are obtained in separate ways, making their vector-space inconsistent. In order to fuse the KG into BART, similar to BART, {\mn} is trained by corrupting texts and then optimizing a reconstruction loss, the cross-entropy, between the decoder’s output and the original texts. We randomly select five concept nodes from our selected entities and mask some concepts among them. {\mn} still takes the entity and relation embedding of all concepts without considering whether the token is masked. Since the graph in the decoder only contains the concept set entities, the decoder is modified as without updating the concept nodes with their neighboring nodes in the pre-training stage. {\mn} is pre-trained to generate the original concept token from the masked concept nodes. For example, ``\textit{[mask] wound [mask] teach soldier}'' in the encoder and ``\textit{student wound treat teach soldier}'' in the decoder. The number of the masked token is randomly sampled from 0 to 5.

\section{Experiment and Analysis}
\subsubsection{Dataset}
{\data}~\cite{lincommongen} is a constrained text generation task, which is to explicitly test the ability of machines on commonsense reasoning when generating a text. The dataset released in this task is constructed through a combination of crowdsourced and existing caption corpora, which consists of 77k commonsense descriptions over 35k unique concept sets. In average, each concept set is composed of 3$\sim$5 unique concepts.
We present the basic statistics of this dataset in Table~\ref{dataset}. Notably, all pairs of concepts in every test concept set are unseen in training data so that it poses a challenge for text generalization. 

\subsubsection{Baselines} 
We compare the performance of our proposed model with several state-of-the-art pre-trained text generation models. 
\textbf{GPT-2}~\cite{radford2019language} is an unidirectional model to predict tokens given the input text in an auto-regressive manner. 
\textbf{UniLM}~\cite{dong2019unified} proposes a unified model of language understanding and language generation using the masked language modeling.
\textbf{UniLM2}~\cite{bao2020unilmv2} further proposes a pseudo-masked language model to learn intra-relations between masked spans via partially auto-regressive modeling. 
\textbf{BERT-Gen}~\cite{bao2020unilmv2} fine-tunes BERT for sequence-to-sequence language generation using a similar training objective employed by UniLM. 
\textbf{T5}~\cite{raffel2019exploring} introduces a unified framework that converts all text-based language problems into a text-to-text format. 
\textbf{BART}~\cite{lewis2020bart} introduces a denoising autoencoder for pre-training sequence-to-sequence models.
For the implementation of those models for the generative commonsense reasoning task, we refer readers to~\cite{lincommongen} for more details. 
\begin{table}[t!]
\centering
\resizebox{0.45\textwidth}{!}{
\begin{tabular}{l|ccc} \toprule
                          & \textbf{Train}  & \textbf{Dev}     & \textbf{Test}     \\ \hline
\textbf{$\#$ Concept sets}            & 32,651 & 993     & 1,497    \\
\textbf{$\#$ Sentences}              & 67,389 & 4,018   & 6,042    \\ \hline \hline
\textbf{$\%$ Unseen Concepts}        & -      & 6.53\%  & 8.97\%   \\
\textbf{$\%$ Unseen Concept-Paris}   & -      & 96.31\% & 100.00\% \\ 
\textbf{$\%$ Unseen Concept-Triples} & -      & 99.60\% & 100.00\% \\ \bottomrule
\end{tabular}
}
\caption{The basic statistics of the {\data} dataset.}
\label{dataset}
\end{table}

\begin{table*}[tp]
\centering
\resizebox{0.8\textwidth}{!}{
\begin{tabular}{l|cc|cc|c|c|c|c}
\bottomrule
\textbf{Model\textbackslash{}Metrics}  & \multicolumn{2}{c|}{\textbf{BLEU-3/4}}& \multicolumn{2}{c|}{\textbf{ROUGE-2/L}} & \textbf{METEOR} & \textbf{CIDEr} & \textbf{SPICE} & \textbf{Coverage} \\
\toprule
\textbf{GPT-2}  ~\cite{radford2019language}    & 30.70 & 21.10 & 17.18  & 39.28 & 26.20  & 12.15 & 25.90 & 79.09    \\
\textbf{BERT-Gen}  ~\cite{bao2020unilmv2}   & 30.40  & 21.10 & 18.05 & 40.49& 27.30  & 12.49 & 27.30 & 86.06    \\
\textbf{UniLM}  ~\cite{dong2019unified}  & 38.30  & 27.70 & 21.48 & \underline{43.87} & 29.70  & 14.85 & 30.20 & 89.19    \\
\textbf{UniLM-v2}~\cite{bao2020unilmv2}   & 31.30 & 22.10  & 18.24 & 40.62 & 28.10  & 13.10 & 28.10 & 89.13    \\
\textbf{T5-Base} ~\cite{raffel2019exploring}  & 26.00  & 16.40 & 14.57 & 34.55 & 23.00  & 9.16  & 22.00 & 76.67    \\
\textbf{T5-Large} ~\cite{raffel2019exploring}   & \underline{39.00}  & \underline{28.60} & 22.01 & 42.97 & 30.10  & \underline{14.96} & \underline{31.60} & 95.29    \\
\textbf{BART} ~\cite{lewis2020bart}  & 36.30 & 26.30 & \underline{22.23}  & 41.98  & \underline{30.90}  & 13.92 & 30.60 & \underline{97.35}  \\ \hline
\textbf{Human Performance}    & 48.20 & 44.90 & 48.88  & 63.79 & 36.20  & 43.53 & 63.50 & 99.31   \\ \hline
\textbf{{\mn}}   &  \textbf{42.10}  & \textbf{30.90} &  \textbf{23.38}  &  \textbf{44.54} &  \textbf{32.40} &  \textbf{16.83}  &  \textbf{32.70}  & \textbf{98.68}\\ \bottomrule
\end{tabular}
}
\caption{Experimental results of different baseline methods on the {\data} test dataset.  We show the best results in boldface, and those with the second best performance are underlined.}
\label{commongen:test}
\end{table*}

\subsubsection{Automatic Evaluation}
Following other conventional generation tasks, we use several widely-used automatic metrics to automatically assess the performance, such as BLEU~\cite{papineni2002bleu}, ROUGE~\cite{lin2004rouge} and METEOR~\cite{banerjee2005meteor}, which mainly focus on measuring n-gram similarities. 
We report the Coverage of concept, which is the average percentage of input concepts that are present after lemmatization. 
In addition, we use evaluation metrics specially designed for image captioning task, such as CIDEr~\cite{vedantam2015cider} and SPICE~\cite{anderson2016spice}. 
These metrics focus on evaluating the associations between mentioned concepts instead of n-gram overlap. 
For example, the SPICE metric uses dependency parse trees as a proxy of scene graphs to measure the similarity of scenarios. 
To estimate human performance within each metric, we treat each reference sentence in test dataset as a \textit{system prediction} and compare it with other references.
It is equivalent to compute inter-annotator agreement. 

Table~\ref{commongen:test} presents the experimental results in a variety of metrics and methods reported on the Leaderboard.\footnote{https://inklab.usc.edu/CommonGen/leaderboard.html}
We can see that {\mn} performs best among all the pre-trained models. 
{\mn} outperforms $7.95\%$/~$8.04\%$ on BLEU-3/4 than the second best model T5-large. 
{\mn} gains 1.15 improvements than the second best model BART on ROUGE-2, the gain 0.67 than UniLM on ROUGE-L.
{\mn} gains 1.50 on METEOR than the second best model BART. 
{\mn} beats the second best model T5-large by $12.50\%$ on CIDEr and $3.48\%$ on SPICE. 
Moreover, {\mn} gets the highest Coverage 98.68 among all baseline pre-trained models. 
The results suggest that leveraging the pre-trained generation model with the knowledge graph can improve the performance of generative commonsense reasoning.

\subsubsection{Human Evaluation}
The automatic evaluations are unable to measure the coherence of the generated text properly.
Therefore, we also access system performance by human evaluation. 
We randomly select 100 instances from the CommonGen test set and invite 3 annotators to access the outputs of different models independently.
Annotators access the overall quality of generative commonsense sentence by ranking them from 1 (worst) to 5 (best) taking into account the following four criteria:
(1) Rationality: is the sentence the reasonable commonsense scenario?
(2) Fluency: is the sentence fluent and grammatical? 
(3) Succinctness: does the sentence avoid repeating information? 
(4) Naturalness: does the sentence use adjunct words? 
The rating of each system is computed by averaging the scores on all test instances.

Table~\ref{human} summarizes the comparison results of five methods. 
Both the percentage of ranking results and overall ratings are reported. 
The results demonstrate that {\mn} is able to generate higher quality output than other models.
Specifically, the outputs generated by {\mn} usually contains more reasonable scenario and are more fluent and precise than other models. 
The human evaluation results further validate the effectiveness of our proposed model.
Moreover, based on the 100 final scores for each approach, we conduct Wilcoxon signed-rank tests~\cite{wilcoxon1970critical}. 
Comparing {\mn} with T5-Large and BART, the $p$-values of Wilcoxon signed-rank testing at $95\%$ confidence level are 1.2$e{-4}$ and 2.9$e{-3}$, which mean the improvements achieved by our approach are statistically significant. 

\begin{table}[tp]
\centering
\resizebox{0.41\textwidth}{!}{
\begin{tabular}{l|cccccc}
\toprule
\textbf{Model} & \textbf{1} & \textbf{2} & \textbf{3} & \textbf{4} & \textbf{5} & \textbf{Rating} \\ \hline
\textbf{GPT-2}  & 22$\%$  & 16$\%$  & 23$\%$  & 20$\%$  & 19$\%$  &  2.98     \\
\textbf{UniLM}  & 5$\%$  & 17$\%$  &  22$\%$ & 24$\%$  & 32$\%$  &    3.61    \\
\textbf{T5-large} &  2$\%$  & 15$\%$  & 12$\%$  & 32$\%$  &  39$\%$   & 3.91       \\
\textbf{BART}      & 1$\%$  & 10$\%$  &  17$\%$ & 30$\%$  &  42$\%$  &  4.02      \\ \hline
\textbf{{\mn}}      & 0 $\%$ & 8$\%$  &  12$\%$ & 25$\%$  & 55$\%$  & $\mathbf{4.27}$  \\
\bottomrule
\end{tabular}
}
\caption{Ranking results of system outputs by human evaluation. 1 is the worst and 5 is the best. The larger rating denotes a better summary quality.}
\label{human}
\end{table}

\begin{figure}[t]
\centering
\includegraphics[width=0.9\linewidth]{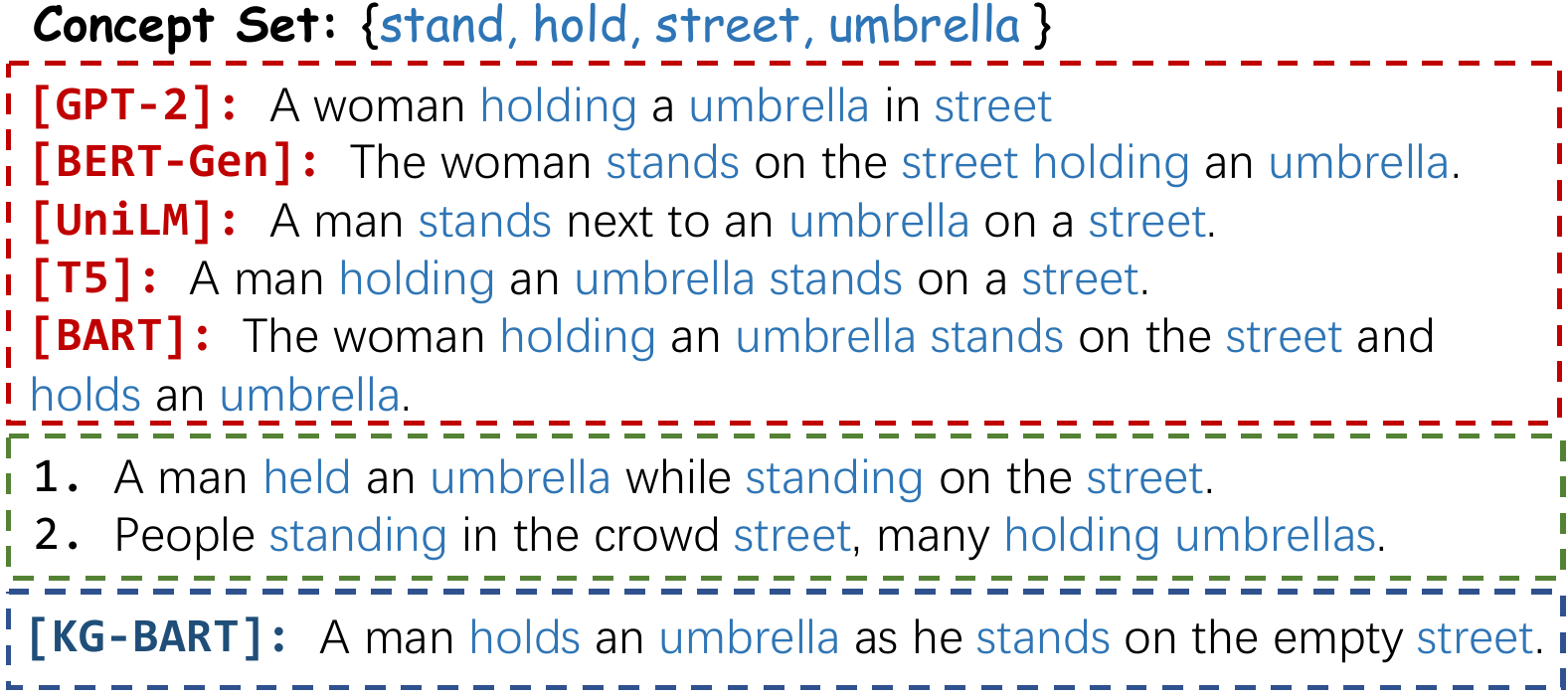}
\caption{A case study of a specific concept set $\{$\textit{stand}, \textit{hold}, \textit{street}, \textit{umbrella}$\}$ for qualitative analysis of machine generations. Human references are collected from AMT.}
\label{figure:case}
\end{figure}

\subsubsection{Case Study}
Figure~\ref{figure:case} gives a specific input concept set $\{$\textit{stand, hold, street, umbrella}$\}$, together with the text generations of different models and human references.
We find that the outputs of fine-tuned pre-trained language models have several problems: 
(1) not covering all concepts, e.g., GPT-2 only covers ``\textit{hold}, \textit{umbrella}, \textit{street}'', ignoring the ``\textit{stand}'', 
(2) unreasonable commonsense relationship between concepts, e.g. in UniLM, the output ``\textit{A man stands next to an umbrella on a street}'' is a rare scenario in daily life,  
and (3) repeating the same content and incorrect grammar, e.g. in BART, it uses both ``\textit{holding an umbrella}'' and ``\textit{holds an umbrella}'', which is repeated information, and in GPT-2, the indefinite article of ``\textit{umbrella}'' should be ``\textit{an}'' rather than ``\textit{a}''. 
By contrast, the output generated by {\mn} covers all concepts and is a relatively reasonable scenario and is comparatively as natural and plausible as the references stated by human.  

\begin{figure}[t]
\centering
\includegraphics[width=1\linewidth]{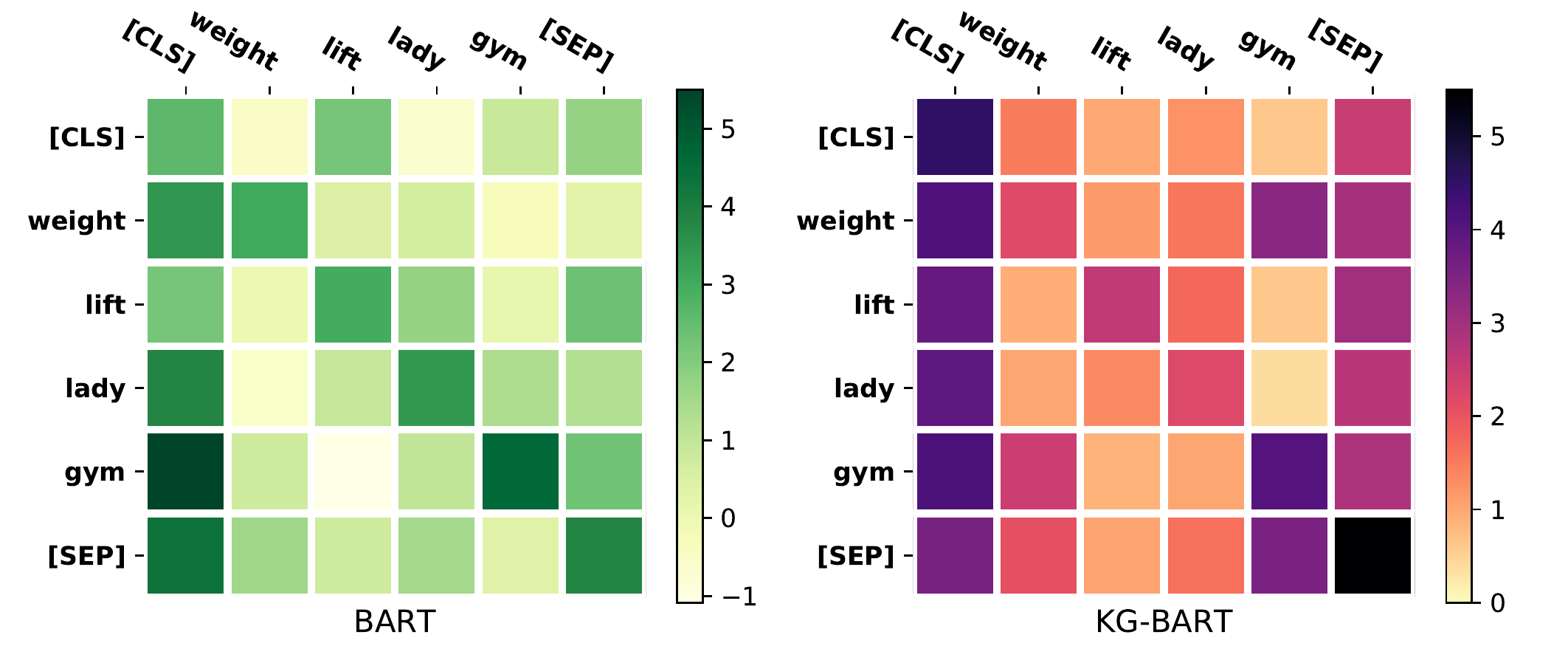}
\caption{Attention weights of the last layers of BART and KG-BART encoder.}
\label{figure:att}
\end{figure}

We also visualize the attention weights of the last layers of KG-BART and BART encoder to validate that our model can capture the better relationship between concepts, as shown in Figures~\ref{figure:att}. 
We can see that the related concept pairs in {\mn} attend much more attention, which is consistent with that in the knowledge graph. 
For example, in practice, ``\textit{weight}'' has a strong relationship with ``\textit{gym}'' on the knowledge graph and the attention weight between them should be large. 
However, this strong relationship has not been demonstrated in BART without knowledge graph.
Therefore, it is reasonable to introduce a knowledge graph as relationship augmentation for better concept representation, also as a guidance to generate more reasonable sentences further.

\subsubsection{Ablation Study}
To evaluate the contributions of individual components of our proposed framework, 
we conduct ablation analysis to investigate the following research questions: 
(1) whether the KG-augmented encoder and decoder improves the performance? (2) whether KG-BART is good at incorporating entity embedding with Transformer? (3) does the KG-BART pre-training works? 

To this end, we test on the following ablations: (1) textual Transformer with only KG-augmented encoder; (2) using the same entity representation at each subword position rather than using SCI and CSD; (3) concatenate the entity embedding with word embedding rather than using MGAT and MHGAT; and (4) without the KG-BART pre-training. 
Table~\ref{table:ab_study} summarizes the ablation results.
It shows that KG-BART can still outperform all these four variants, certifying the effectiveness of each designed component in our model and we can also see that incorporating KG with the pre-trained model can help the model achieve a better performance. 
\begin{table}[tp]
\centering
\resizebox{0.45\textwidth}{!}{
\begin{tabular}{llll}
\toprule
\multicolumn{2}{l}{Ablation methods}   & BLEU-3/4 & ROUGE-2/L \\ \hline
(1) KG-Aug Enc. \ding{51} & Dec. \ding{55} &   40.40/29.40   &    22.66/43.13    \\
(2) SCI \ding{55} & CSD \ding{55} & 41.20/29.70 & 23.15/43.57 \\
(3) MGAT \ding{55} & MHGAT \ding{55} & 40.90/29.30 & 22.96/43.78 \\
\multicolumn{2}{l}{(4) Pre-training\ding{55}}   &  39.80/27.90    &   21.87/42.92    \\
\bottomrule
\end{tabular}
}
\caption{Ablation study of the proposed model. SCI, CSD, MGAT and MHGAT are KG-BART components.}
\label{table:ab_study}
\end{table}
\subsubsection{Transfer {\mn} to Commonsense QA}
\begin{figure}[t]
\centering
\includegraphics[width=0.72\linewidth]{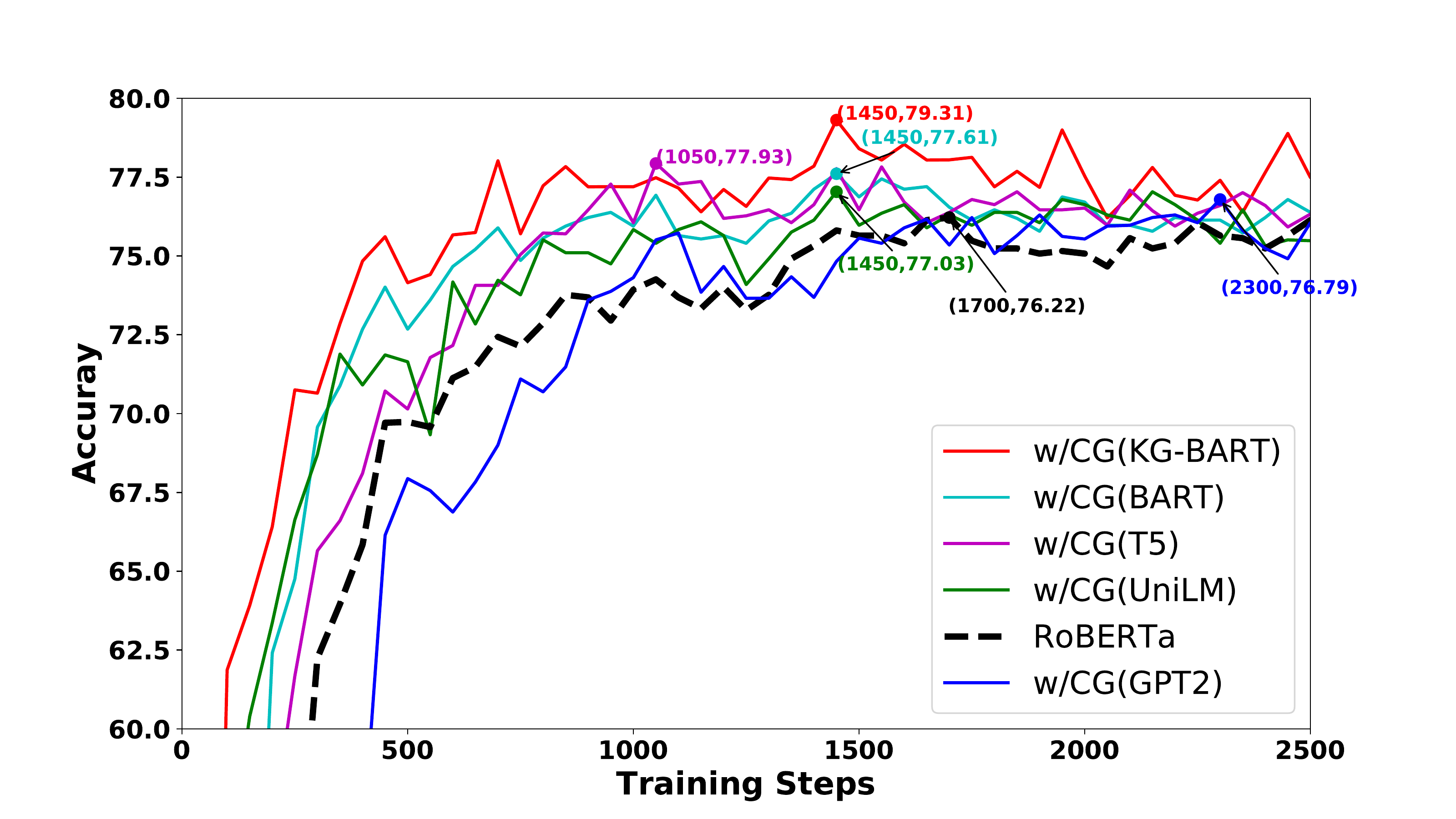}
\caption{The learning curve of transfer study on CSQA.}
\label{figure:tranf}
\end{figure}

We also investigate whether the ability of generative commonsense reasoning in {\mn} can benefit commonsense-centric downstream tasks such as Commonsense Question Answering~(CSQA)~\cite{talmor2018commonsenseqa}.
We use the models trained on the {\data} dataset for generating useful context to the question. 
We extract the \textit{nouns} and \textit{verbs} in questions and five choices, and combine the concepts of question $q$ and each choice $c_i$ to build concept sets. 
Then, we construct the concept-reasoning and concept-expanding graphs based on concepts and use these concept sets and the graphs as inputs to {\mn} to generate the context sentence $g_{i}$ for each choice. 
Finally, we prepend the outputs in front of questions, i.e., ``\textit{$<$s$>$G:$\text{g}_{i}$ $<$/s$>$ Q:q $<$/s$>$ C:$\text{c}_{i}$ $<$/s$>$}''. 
The RoBERTa~\cite{liu2019roberta} model for CSQA uses the same form without ``\textit{G:$\text{g}_{i}$ $<$/s$>$}'' in fine-tuning. 
We show the learning curve in Figure~\ref{figure:tranf}, where $X$ axis is the number of training steps and $Y$ axis is the accuracy on official dev dataset.

We find that in most cases, using the context generated by pre-trained models can further improve the performance of original RoBERTa by a large margin. 
Especially, {\mn} converges at better accuracy from 76.22 (in original RoBERTa) to 79.31 and it outperforms other baselines. 
We find that the context generated by our model {\mn} can speed up training about 2.5 times, if we look at the 550th steps of {\mn} (75.51) and 1,400th steps of original RoBERTa (75.31). 
Note that in the beginning training steps, GPT-2 causes negative transfer due to the low quality of generated context.
Through manual analysis, we find that {\mn} can generate more reasonable and natural sentences for correct choices while noisy sentences for wrong choices. 
For example, $q$=``\textit{What would you do if you \underline{want} to be able to \underline{earn} \underline{money}?}'', $c_{i}$=``\textit{\underline{apply} \underline{for} \underline{job}}'' (correct) with $g_i$=``\textit{applying for a job so i would earn money.}''; $c_j$=``\textit{\underline{stand} \underline{in} \underline{line}}'' (wrong) $g_j$=``\textit{i would want to earn money standing in line to get a deal on a product.}'' 
\section{Related Work}

\subsubsection{Incorporating Commonsense for NLG}
There are a few recent works that incorporate commonsense knowledge in language generation tasks such as storytelling~\cite{guan2019story}, visual storytelling~\cite{yang2019knowledgeable}, essay generation~\cite{yang2019enhancing}, image captioning~\cite{lu2018neural}, evidence generation~\cite{liu2020commonsense} and conversational generation systems~\cite{zhang2020grounded}. These works suggest that generative commonsense reasoning has great potential to benefit downstream applications. Our proposed model {\mn}, to the best of our knowledge, is the first work on equipping the pre-trained language generation model with the external commonsense knowledge for the constrained language generation.

\subsubsection{Enhancing Pre-Trained Model with Knowledge}
Recently, several works have attempted to learn joint representation learning of words and entities for effectively leveraging external KGs on language understanding tasks and achieved promising results. ERNIE~\cite{zhang2019ernie} incorporates informative entities from KG aligning with context to enhance pre-training language understanding. KEPLER~\cite{wang2020kepler} encodes textual descriptions of entities with a pre-trained language understanding model, and then jointly optimize the knowledge embedding and language modeling objectives. K-BERT~\cite{liu2020k} injects domain knowledge into the models by adding triples from the knowledge graph as supplementary words. Inspired by these works, we argue that extra knowledge information can effectively benefit existing pre-training models on the language understanding tasks. In this paper, we utilize KGs to train an enhanced language generation model by incorporating the entity relationships to improve the language representation. 

\section{Conclusion}
We have presented a KG-augmented approach \mn based on pre-trained BART for generative commonsense reasoning. 
Through capturing the relations among concepts over a KG, \mn can generate high-quality sentences even in the unseen concept sets. 
{\mn} further considers the neighbor entities of each concept node as to generate more natural and logical sentences. It can also be extended to any seq2seq pre-trained language generation models, like T5~\cite{raffel2019exploring} and MASS~\cite{song2019mass}.
Experimental results demonstrate that {\mn} has better abilities of both commonsense reasoning and text generalization.

\section{Broader Impact}
This paper has proposed to incorporate the KG into the pre-trained language generation model to produce a more natural and plausible output for generative commonsense reasoning task. 
The proposed approach {\mn} offers a new way to improve the quality of pre-trained language generation with KG and will potentially be applied in other language generation tasks, e.g., text summarization, dialogue response generation and abstractive QA. Specifically, our approach has the following potential impacts.

Our approach is designed to improve the pre-trained language generation models with a novel KG-Augmented mechanism to capture KG structure and semantic information. From the experimental evaluations, we can find that this mechanism has the capacity of improving system performances by a significant margin. Hence, this work may inspire the researchers from the community of language generation by injecting external knowledge to enrich semantic representation. 
Specifically, rather than using commonsense KG ConceptNet, our model is easy to be extended to some other KG datasets such as encyclopedia KG databases, Freebase~\cite{bollacker2008freebase} and DBpedia~\cite{auer2007dbpedia} to enrich the semantic representation by capturing the relationships between the mentioned entities in context and generate more detailed output by considering the additional useful information from KG of the mentioned entities.

On the other hand, the goal of conversational Artificial Intelligence (AI) is to create intelligent systems that can simulate human-like thinking and reasoning process. 
To the best of our knowledge, all the current data-driven conversational agents like Apple's Siri, Google Assistant and Amazon's Alexa are struggling at achieving the ability of commonsense reasoning on generating the human-like responses.
However, one distinguishing characteristic of our approach {\mn} is that it can use automated commonsense reasoning to truly ``\textit{understand}'' the context and provide rational and plausible responses as natural as possible. Thus by adapting {\mn} to dialogue response generation, we believe that it will significantly boost the generative commonsense reasoning capability and benefit real-world applications in the conversational AI systems.

\section{Acknowledgement}
We would like to thank all the reviewers for their helpful comments.
This work is supported by NSF under grants III-1763325, III-1909323, and SaTC-1930941.

\clearpage

\begin{appendices}
\section{Appendix}
\subsection{Training Details and Parameters}
To implement the TranE model for KG embedding, we use the open source OpenKE,\footnote{https://github.com/thunlp/OpenKE} and dimension of entity embedding $d_e$ and relation embedding $d_r$ to 1,024. The quantity of the select concepts for training TransE is 12K and covers all concept entities in {\data}.
In the pre-training procedure of \mn, we sample 200K five-concept sets from those select concepts. 
The entity embeddings and relation embeddings are fixed during pre-training. Since the pre-training is computation costly, we start pre-training from BART's released checkpoint and randomly initialize KG-Augmented Transformer in {\mn} with $\mathcal{N}(0,0.02)$. 
We further train {\mn} for 0.2 million steps on a Nvidia Titan-RTX 24GB GPUs. 

Our implementation of {\mn} is based on BART code,\footnote{https://github.com/huggingface/transformers} which is implemented based on PyTorch. 
In detail, we have the following model size: the layer number of Textual Transformer $N=6$, the layer number of KG-Augmented Transformer $M=6$, the dimension of token embedding $d_{w}=1024$, multi-heads $K=16$ and the kernel size $l$ of CNN is set to 2. 
We tokenize the text using the byte-pair encoding same as GPT-2~\cite{radford2019language}, with the maximum length of 32 for encoder and 64 for decoder.  
We used AdamW~\cite{loshchilov2019decoupled} with $\beta_1 = 0.9$, $\beta_2 = 0.98$, and $\epsilon = 1e-6$ for optimization. 
We set the initial learning rate from $\{8e-6, 1e-5, 2e-5, 3e-5\}$ with warm-up rate of 0.1 and $L_2$ weight decay of 0.01. 
The batch size is selected from $\{$16, 24, 32$\}$. We employ half-precision training (floating points 16) using \texttt{apex}\footnote{https://github.com/NVIDIA/apex} to reduce memory consumption and speed-up training. 
We train all models with maximum likelihood estimation, and use label smoothing~\cite{szegedy2016rethinking} with smoothing factor 0.1. 
In the fine-tuning process, the model is trained with a maximum number of 5 epochs and the gradients are accumulated every four steps.
We apply dropout with probability 0.1 to avoid over-fitting. 
During inference, we use beam search with beam size 5 and length penalty with factor 0.6.

\subsection{Baseline Implementation}
GPT-2 for this sequence-to-sequence task, we condition the language model on the format ``\textit{$c_1 c_2 \cdots c_k = y$}'' during fine-tuning, where $c_i$ is a concept in the given concept-set and connects with other concepts with a blank; y is a target sentence. For inference, we sample from the fine-tuned GPT-2 model after a prompt of ``\textit{$c_1 c_2 \cdots c_k = y$}'' with beam search and use the first generated sentence as the output sentence. For BERT-Gen, we use the s2s-ft package3 to finetune them in a sequence-to-sequence fashion that is similar to the LM objective employed by UniLM.

As for T5, the state-of-the-art text-to-text pre-trained model which is pre-trained with a multitask objective by extending a task description before the input text, we extend the input concept set with a simple prompt: ``\textit{generate a sentence with:}'' and fine-tune the model with the source sentence on the format ``\textit{generate a sentence with $c_1 c_2 \cdots c_k = $}'' For decoding, we employ the standard beam search with a beam size of 5 for all compared models.

\subsection{Error Analysis}
We investigate the error cases found by examining the generated sentences with low evaluation scores and find three types of errors:

The first error is that our KG-BART tends to generate a long sentence to cover the concept set. For example, given a concept set “\textit{\{talk phone wear\}}”, our KG-BART will generate “\textit{A man and a woman are talking on the phone and one of them is wearing glasses.}”, while the human ground truth is “\textit{A man wearing glasses is talking on a phone.}” 

The second error of our model is that KG-BART suffers repeatedly generating the same concept. For example, given the concept set “\textit{\{roll ball lane pin\}}”, our KG-BART generates “\textit{A man rolls a bowling ball down a bowling lane and pins the ball down the lane.}”, while the human ground truth is “\textit{The bowling ball rolled straight down the center of the lane and knocked down all of the pins.}”

The third error of our model is that in some sentences, the generated sentences by KG-BART are still different from human commonsense. For example, given a concept set “\textit{\{jump water cliff watch\}}”, our KG-BART will generate “\textit{The boy jumped off the cliff to watch the water.
}”, while the human ground truth is “\textit{Watch him jumping from the cliff to the water.}”

We think those limitation caused by KG-BART can only learn well the local relation between each concept pair, which means learning the pattern between concept pairs, so the proposed method is good at generating correct phrase. But it fail to capture the global relationship which need to check the relation between phrase in the generated sentence. How to capture this global relationship will be our future work direction.

\end{appendices}

\end{document}